\documentclass[final]{cvpr}

\usepackage{times}
\usepackage{epsfig}
\usepackage{multirow}
\usepackage{graphicx}
\usepackage{amsmath}
\usepackage{amssymb}
\usepackage[pagebackref=true,breaklinks=true,colorlinks,bookmarks=false]{hyperref}
\usepackage{xcolor}
\definecolor{applegreen}{rgb}{0.55, 0.71, 0.0}
\definecolor{cadmiumgreen}{rgb}{0.0, 0.42, 0.24}
\definecolor{brinkpink}{rgb}{0.98, 0.38, 0.5}
\definecolor{davy\'sgrey}{rgb}{0.33, 0.33, 0.33}
\definecolor{darkslategray}{rgb}{0.18, 0.31, 0.31}
\usepackage[sort,nocompress]{cite}

\def\cvprPaperID{10700} 
\def\confYear{CVPR 2021}

\begin{document}

\title{Hallucination Improves Few-Shot Object Detection}

\author{Weilin Zhang
\qquad
Yu-Xiong Wang\\
University of Illinois at Urbana-Champaign \\ 
{\tt\small \{weilinz2, yxw\}@illinois.edu}
}

\maketitle
\thispagestyle{empty}

\begin{abstract}
Learning to detect novel objects from few annotated examples is of great practical importance. A particularly challenging yet common regime occurs when there are extremely limited examples (less than three). One critical factor in improving few-shot detection is to address the lack of variation in training data. We propose to build a better model of variation for novel classes by transferring the shared within-class variation from base classes. To this end, we introduce a hallucinator network that learns to generate additional, useful training examples in the region of interest (RoI) feature space, and incorporate it into a modern object detection model. Our approach yields significant performance improvements on two state-of-the-art few-shot detectors with different proposal generation procedures. In particular, we achieve new state of the art in the extremely-few-shot regime on the challenging COCO benchmark. 
\end{abstract}

\section{Introduction}
\label{sec:intro}
Modern deep convolutional neural networks (CNNs) rely heavily on large amounts of annotated images~\cite{russakovsky2015imagenet}. This data-hungry nature limits their applicability to some practical scenarios such as autonomous driving, where the cost of annotating examples is prohibitive, or which involve {\em never-before-seen} concepts~\cite{zhu2014capturing,fink2011acquiring}. By contrast, humans can rapidly grasp a new concept and make meaningful generalizations, even from a single example~\cite{Schmidt2009}. To bridge this gap, there has been a recent resurgence of interest in few-shot or low-shot learning that aims to learn novel concepts from very few labeled examples~\cite{FeiFeiTPAMI2006,VinyalsNIPS2016,WangECCV2016,snell2017prototypical,FinnICML2017}.
\begin{figure}[h]
    \centering
    {\includegraphics[width=8cm]{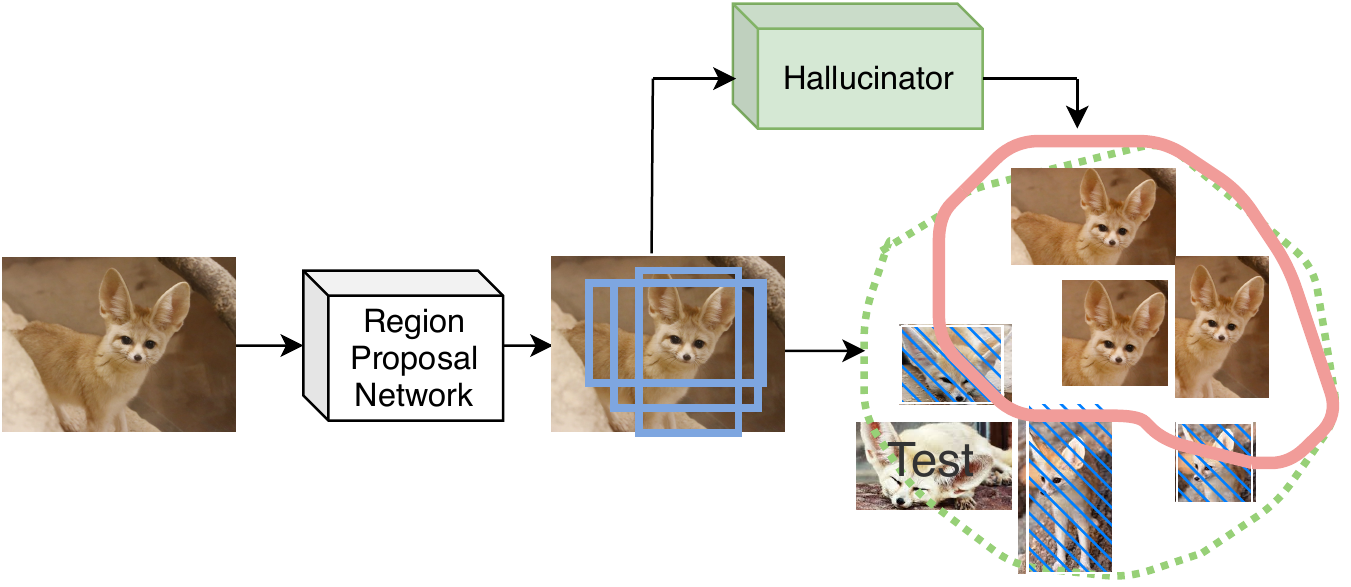}
    \caption{Learning to detect a novel class, {\em fennec fox}, from a single training example (\ie, 1-shot detection) using a serial detector. The region proposal network (RPN) generates a few high intersection-over-union (IoU) boxes for the detector's classifier. The \textcolor{brinkpink}{pink} circle represents the classifier decision boundary learned from these boxes. Due to a lack of sample variation, the decision boundary is not accurately estimated. With hallucinated examples (image in backslash) produced by our hallucinator, the classifier learns a better decision boundary (\textcolor{cadmiumgreen}{the dotted circle}), thus being able to potentially correct previously misclassified instances.}
    \label{fig:example}}
\end{figure}

Despite notable successes, most of the existing work has focused on simple classification tasks with artificial settings and small-scale datasets~\cite{VinyalsNIPS2016,snell2017prototypical}. However, few-shot object detection, a task of great practical importance that learns an object detector from only a few annotated bounding box examples~\cite{kang2018few,WangRH19,wang2020few}, is far less explored. Few-shot detection requires determining {\em where} an object is as well as {\em what} it is (and handling distracting background regions~\cite{girshick2014rich}, \etc), and is much harder than few-shot classification. The most difficult regime occurs when there are {\em very limited examples} (less than 3) for novel classes (Figure~\ref{fig:example}), which is a common yet extremely challenging case in the real world.

While few-shot classification approaches are helpful (\eg,~\cite{wang2015model,chen2018lstd,dong2018few,kang2018few,schwartz2018repmet}), few-shot detection is much more than a straightforward application of few-shot classification approaches. The state-of-the-art two-stage fine-tuning approach (TFA)~\cite{wang2020few} learns a better representation for few-shot detection, through (1) pre-training on base classes with abundant data and then only fine-tuning the box classifier and regressor on novel classes, and (2) introducing instance-level feature normalization to the box classifier during fine-tuning. Despite the improvement, its performance in the extremely low-data regime is still far from satisfying. 

We argue that, to fully improve {\em extremely-few-shot} detection performance, a key factor is to effectively deal with {\em the lack of variation in training data}. This is because for an object detector to be accurate, its classifier must build a useful model of variation in appearance with very few examples. More concretely, a modern object detector first finds promising image locations, typically boxes, using a region proposal network (RPN)~\cite{ren2015faster}, then passes promising boxes through a classifier to determine what object is present, and finally performs various cleanup operations such as non-maximum suppression (NMS), aimed at avoiding duplicate predictions and improving localization. Now assume that the detector must learn to detect a novel category from a single example (Figure~\ref{fig:example}). The only way the classifier can build a model of the category's variation in appearance is by learning from the high intersection-over-union (IoU) boxes reported by the RPN. Although there is variation of boxes produced by the RPN, the variation from a single example is too weak to train the classifier for the novel class.

To overcome this issue, one strategy is to adjust the learning procedure for RPN, so that it reports highly informative boxes. Contemporary work~\cite{CoRPNs} achieves this by training multiple RPN's be somewhat redundant and cooperating. Hence, if one RPN misses a highly informative box, another will get it. This cooperating RPN's (CoRPNs) approach, while helpful, is still insufficient. In the extremely-few-shot regime, all positive novel class proposals produced by the multiple RPN's are only slightly modified from and thus similar to the few available positive instances (with light-weighted cropping and scaling operations); their variation is significantly limited for building a strong classifier.

In this paper, we propose a different perspective on building a model of variation for novel classes, {\em by transferring the shared within-class variation from base classes}. In fact, many modes of variation in the visual world (\eg, camera pose, lighting changes, and even articulation) are shared across categories and can generalize to unseen classes~\cite{salakhutdinov2012one}. While such within-class variation is difficult to be encoded through the proposal generation procedure, it can be effectively captured by {\em learning to hallucinate examples}~\cite{wang2018low}.

To this end, we introduce a {\em hallucinator network} into a modern object detection model. The hallucinator network performs data hallucination for the box classifier in the learned region of interest (RoI) feature space. We train the hallucinator on data-abundant base classes, encoding the rich structure of their shared modes of variation. We then use the learned hallucinator to generate additional novel class examples and thus produce an augmented training set for building better classifiers, as shown in Figure~\ref{fig:example}.

Note that the existing strategy for training the hallucinator in few-shot classification~\cite{wang2018low} is coupled with a complicated meta-learning process, making it difficult to apply to state-of-the-art few-shot detectors like TFA~\cite{wang2020few} or CoRPNs~\cite{CoRPNs}. We overcome this challenge by introducing a much simpler yet effective training procedure: we train our hallucinator and the detector's classifier in an {\em EM-like} (expectation-maximization) manner, where a ``strongest possible'' classifier is trained first with all the available base class data; the hallucinator is then trained under the guidance of this already-trained classifier; and finally, the classifier is re-trained and refined based on the set of augmented examples (with hallucinated examples) on novel classes.



{\bf Our contributions} are three-fold. (1) We investigate a critical yet under-explored issue in extremely-few-shot detection (\eg, as few as one) -- the lack of variation in training data. (2) We propose a novel data hallucination based approach to address this issue, which effectively transfers shared modes of within-class variation from base classes to novel classes. Our approach is simple, general, and can work with different region proposal procedures. (3) Our approach significantly outperforms the state-of-the-art TFA~\cite{wang2020few} and most recent cooperating RPN's~\cite{CoRPNs} detectors in the extremely-few-shot regime. Our code is available at \url{https://github.com/pppplin/HallucFsDet}. 


\section{Related Work}
\label{sec:related}
\textbf{Object Detection:} Modern detectors are typically based on convolutional neural networks with two types of architectures -- serial detection~\cite{girshick2015fast} and parallel detection~\cite{redmon2016you}. Both families run a region proposal process~\cite{EndresHoiem, Regionproposal} that determines whether an image region contains an object or not. They differ in when to run the region proposal process. Serial detectors (or two-stage detectors) first generate promising region proposals and then feed each proposal box to a classifier that predicts if the region contains an object. Serial detectors include R-CNN~\cite{girshick2014rich} and its variants, such as Fast R-CNN~\cite{girshick2015fast}, Faster R-CNN~\cite{ren2015faster}, Mask R-CNN~\cite{he2017mask}, SPP-Net~\cite{he2015spatial}, FPN~\cite{lin2017feature}, and DCN~\cite{dai2017deformable}.  

Parallel detectors (or one-stage detectors) run the region proposal process and the classification process simultaneously. Parallel detectors are usually faster than serial detectors at the expense of decreasing accuracy, since the classifier has no prior knowledge of whether a box contains an object. Representative parallel detectors are variants of YOLO~\cite{redmon2016you, redmon2017yolo9000, redmon2018yolov3, bochkovskiy2020yolov4}, SSD~\cite{liu2016ssd}, CornerNet~\cite{law2018cornernet}, and ExtremeNet~\cite{zhou2019bottom}. Our work introduces a general data hallucination strategy to improve the detection performance in the few-shot regime. While we mainly focus on serial detectors in this paper since they achieve state-of-the-art few-shot detection performance, we believe that our strategy applies to parallel detectors as well, which we leave as future work.


\textbf{Few-Shot Object Detection:} Advanced few-shot detectors are usually built in a serial fashion~\cite{WangRH19, yan2019metarcnn, wang2020few, fan2020fsod, wu2020mpsr, Xiao2020FSDetView}. One line of work focuses on learning better feature representations through metric learning~\cite{schwartz2018repmet, yang2020restoring, li2021maxmargin}, in ways of modeling multi-modal distributions in each class~\cite{schwartz2018repmet}, restoring negative information~\cite{yang2020restoring}, and reserving adequate margin space among novel classes~\cite{li2021maxmargin}. Modified fine-tuning techniques have also been explored. For example, a regularized fine-tuning approach is proposed to transfer knowledge from a pre-trained detector to a few-shot detector~\cite{chen2018lstd}. Recently, a simple two-stage fine-tuning approach has been shown to outperform more sophisticated methods~\cite{wang2020few}. Other work seeks improvements by applying meta-learning techniques, such as learning a meta-model to reweight pre-trained features given few-shot data~\cite{kang2018few} and attaching meta-learned classifiers to Faster R-CNN~\cite{WangRH19,yan2019metarcnn}.

Another line of work focuses on improving the proposal generation process, by introducing attention mechanisms and generating class-aware features for classifiers~\cite{Hsieh19AttenFew, fan2020fsod, Xiao2020FSDetView, osokin20os2d, CoRPNs, sun2021fsce}. Some work modifies the proposal ranking process and ranks proposals based on similarity with query images~\cite{Hsieh19AttenFew, fan2020fsod}. An RPN ensemble method is proposed to avoid missing highly informative proposal boxes~\cite{CoRPNs}. Contrastive-aware object proposal encodings are further learned to reduce the possibility of misclassifying novel class objects to confusable classes~\cite{sun2021fsce}. Additional information has also been shown helpful, such as semantic relations~\cite{zhu2021semantic} and multi-scale representations~\cite{wu2020mpsr}. {\em Orthogonal to existing work}, we address few-shot detection by hallucinating additional data and enriching sample variation.

\textbf{Data Hallucination:} Despite recent progress on learning to hallucinate examples to deal with data scarcity~\cite{hariharan2016low,wang2018low,gao2018low,schwartz2018delta,zhang2018metagan}, much of the work has focused on classification tasks and performed in a learned feature space. Novel class features are generated by learning shared feature transformations from base classes~\cite{hariharan2016low}. Pairwise deformations between examples of the same class are captured and used to generate novel class instances~\cite{schwartz2018delta}. A meta-learner and a hallucinator are combined and jointly optimized to boost recognition performance~\cite{wang2018low}. Our approach builds on the feature space hallucination framework of~\cite{wang2018low}, but makes significant modifications to the hallucinator architecture and the hallucination procedure. To the best of our knowledge, our work is the {\em first} to demonstrate the effectiveness of data hallucination for few-shot detection.

\section{Our Approach}
\label{sec:method}
We believe that hallucination would improve any few-shot detection models with standard architectures. In this section, we focus on improving two state-of-the-art few-shot detectors with {\em different region proposal mechanisms}.

\textbf{Few-Shot Object Detection Setting and Evaluation Procedure:} We follow the few-shot detection setting and the standard evaluation procedure introduced in~\cite{kang2018few, wang2020few}. Classes are split into two sets: base classes $C_\text{b}$ and novel classes $C_\text{n}$. Both models which we study adopt a two-stage fine-tuning procedure~\cite{wang2020few}. A model is trained on base classes in the first stage and is then fine-tuned on novel classes in the second stage. In the first stage, the model only detects base classes -- the model is a $|C_\text{b}|$-way detector. In the second stage, the model is expanded as a $(|C_\text{b}| + |C_\text{n}|)$-way detector. In order to maintain performance on base classes while learning to detect novel class instances, the detector is fine-tuned on a {\em balanced} few-shot dataset containing both novel and base classes. Finally, we compare mean average precision (AP) on novel classes.

\begin{figure}[t!]
\centering\includegraphics[width=\linewidth]{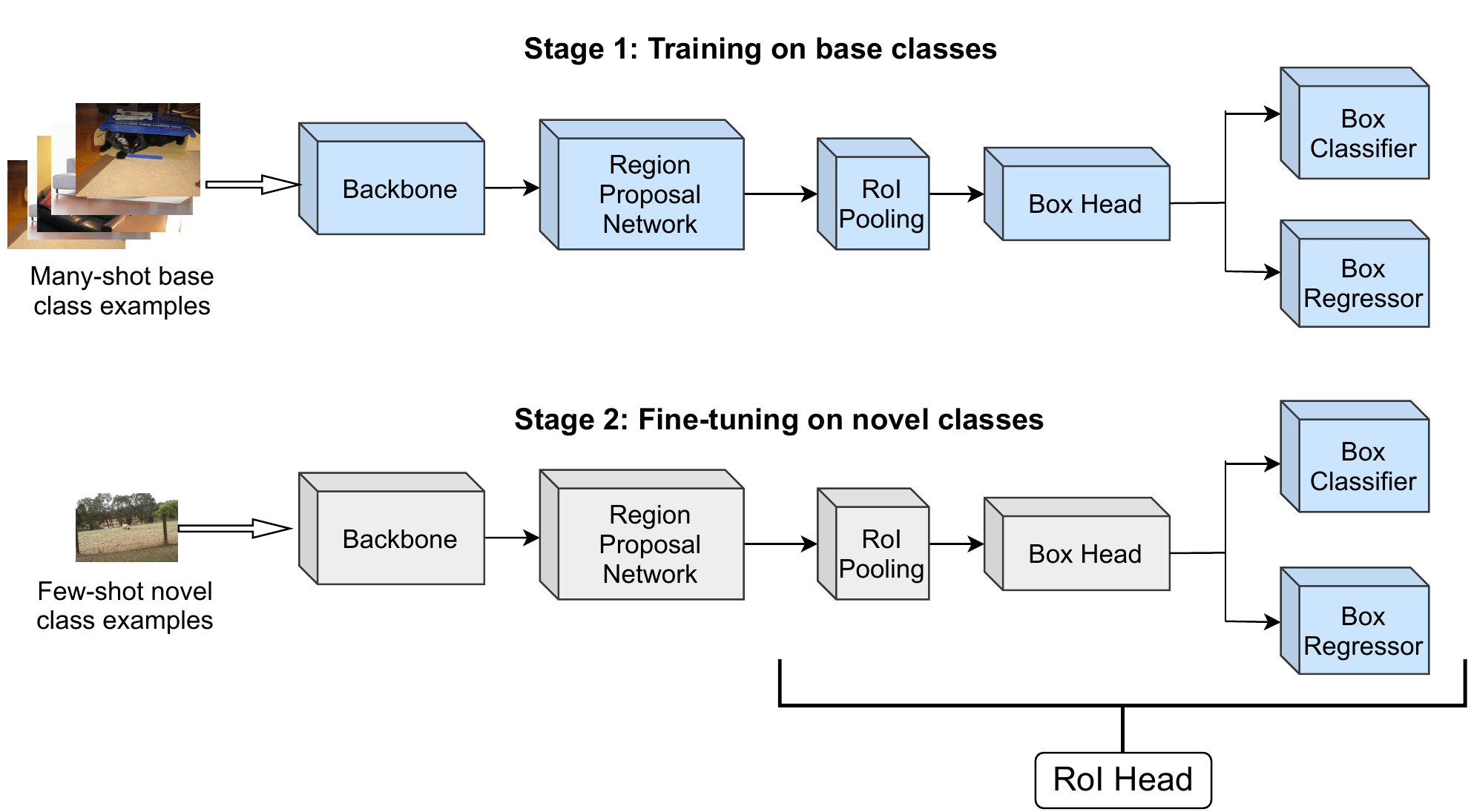}
\caption{Illustration of the two-stage fine-tuning approach (TFA)~\cite{wang2020few} under the Faster R-CNN framework. In stage 1, the whole model is trained with base class images. In stage 2, only the classifier and the bounding box regressor in the RoI head are fine-tuned with novel class images. A module in {\color{blue}blue} is trained during the corresponding stage, while a module in {\color{darkgray}gray} is frozen.}
 \vspace{-4mm}
\label{two_stage}
\end{figure}
\subsection{Few-Shot Object Detection Models}
\label{sec:tfacorps}
\textbf{Two-stage Fine-tuning Approach (TFA):} A two-stage fine-tuning approach is introduced~\cite{wang2020few} under the widely-used Faster R-CNN framework, which significantly improves few-shot detection performance. As a serial detector, Faster R-CNN consists of a backbone, a region proposal network (RPN), a region of interest (RoI) pooling layer, an RoI feature extractor, a bounding box classifier, and a bounding box regressor (Figure~\ref{two_stage}). The last four components together construct the RoI head. The backbone extracts image features, which are passed through the RPN to produce promising areas with potential objects. The RoI head then transforms, classifies, and refines potential object boxes into labeled boxes. TFA modifies the standard Faster R-CNN by using a cosine similarity based classifier to reduce intra-class variance for few-shot learning. 
TFA uses an ImageNet~\cite{russakovsky2015imagenet} pre-trained ResNet-101 with a feature pyramid network~\cite{lin2017feature} as the backbone. As shown in Figure~\ref{two_stage}, in the first stage, TFA trains the whole model using base class images. In the second stage, only the classifier and the bounding box regressor in the RoI head are fine-tuned using novel class instances, with the rest of the model frozen.

\begin{figure}[t!]
\centering
\includegraphics[width=\linewidth]{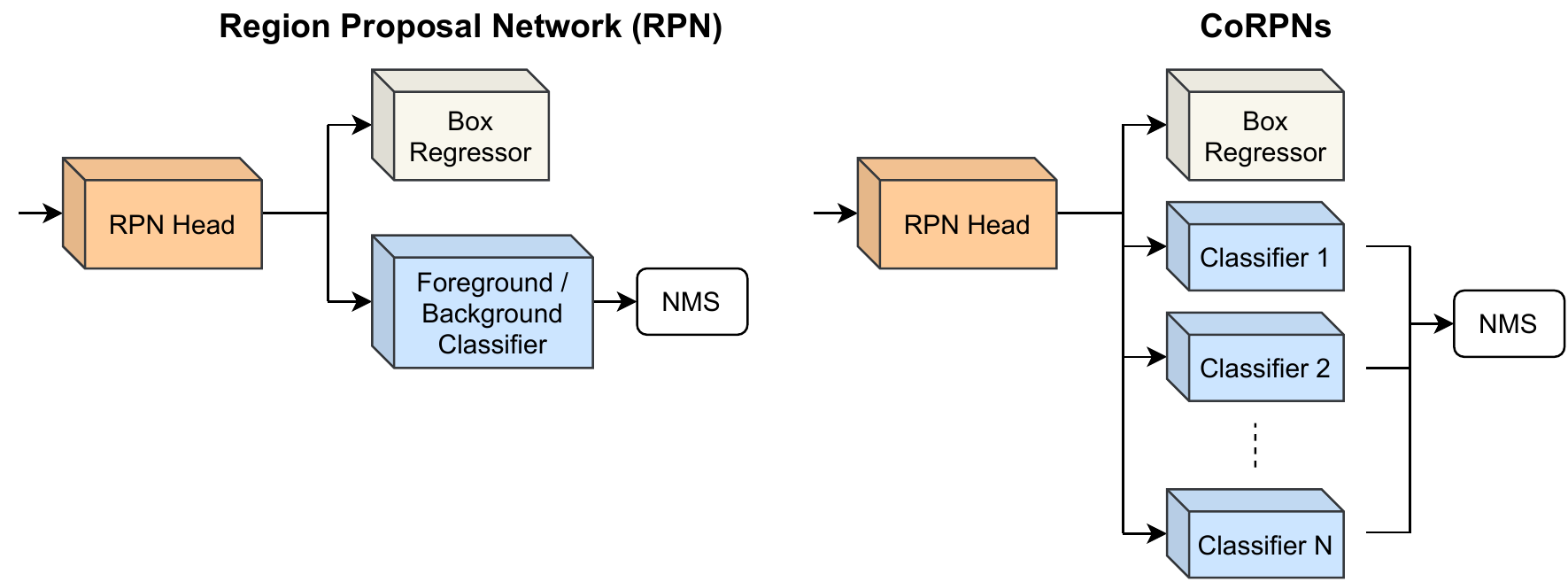}
\caption{Illustration of cooperating RPN's (CoRPNs)~\cite{CoRPNs}. {\bf Left}: the standard RPN structure in Faster R-CNN. {\bf Right}: CoRPNs with $N$ RPN classifiers.}
\vspace{-4mm}
\label{rpn}
\end{figure}

\textbf{Cooperating RPN's (CoRPNs):} 
An improved proposal generation mechanism based on RPN ensemble is proposed in CoRPNs~\cite{CoRPNs}, which produces highly informative proposal boxes for few-shot detection. As shown in Figure~\ref{rpn}, CoRPNs train multiple distinct but cooperating RPN's so that if one RPN misses a highly informative proposal box, another one will get it. Except the proposal generation procedure, CoRPNs share the same model architecture and training strategy with TFA. Specifically, CoRPNs train $N$ RPN's by a modified RPN classification loss

\begin{equation}
\setlength{\abovedisplayskip}{3pt}
\setlength{\belowdisplayskip}{3pt}
\mathcal{L} = \mathcal{L}^{j^*}_\text{CE} + \mathcal{L}_\text{div} + \mathcal{L}_\text{coop},
\end{equation}
where $\mathcal{L}^{j^*}_\text{CE}$ is a binary cross-entropy loss of a selected RPN $j^*$. CoRPNs select the most certain RPN for every box. For instance, each anchor box will get the RPN $j^*$'s score that has a probability closer to either 0 or 1. The selected RPN $j^*$ gets this box's gradient at training. 


The divergence loss $\mathcal{L}_\text{div}$ encourages RPN's to be different, and the cooperation loss $\mathcal{L}_\text{coop}$ enforces RPN's to cooperate by setting a lower bound for every RPN's response to a foreground box. $\mathcal{L}_\text{div}$ is defined as 
\begin{equation}
\setlength{\abovedisplayskip}{3pt}
\setlength{\belowdisplayskip}{3pt}
\mathcal{L}_\text{div} = -\log(\det(\Sigma(\mathcal{F}))),
\end{equation}
which is the negative log of the covariance matrix on every RPN's prediction to proposal boxes. 
For foreground box $i$ and RPN $j$, the cooperation loss is 
\begin{equation}
\setlength{\abovedisplayskip}{3pt}
\setlength{\belowdisplayskip}{3pt}
\mathcal{L}_\text{coop}^{i, j}  =\max\{0, \phi - f^{j}_{i}\},
\end{equation}
where $\phi$ is the lower bound of every RPN's response to a foreground box, and $f^{j}_{i}$ is RPN j's response to the box i. $\mathcal{L}_\text{coop}$ is averaged over all RPN's and foreground boxes.

\subsection{Few-Shot Object Detection with Hallucination}
\begin{figure}[t!]
\centering
\includegraphics[width=.9\linewidth]{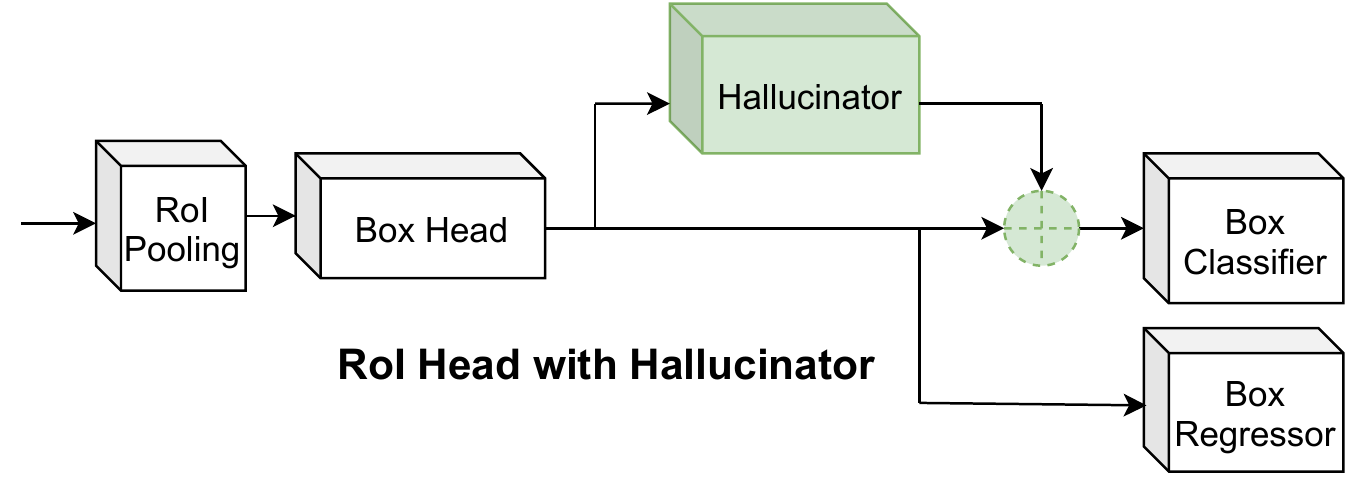}
\caption{Illustration of the RoI head structure with a hallucinator module. The hallucinator is inserted right before the classifier. The hallucinated examples are appended to the original training examples to train the classifier. The bounding box regressor is not affected by the hallucinator.}
\vspace{-4mm}
\label{halluc}
\end{figure}
We introduce a hallucinator network $H$ with parameters $\phi$ that learns to generate additional examples for novel classes by leveraging the shared within-class feature variation from base classes.
\begin{figure}[t!]
\centering
\includegraphics[width=\linewidth]{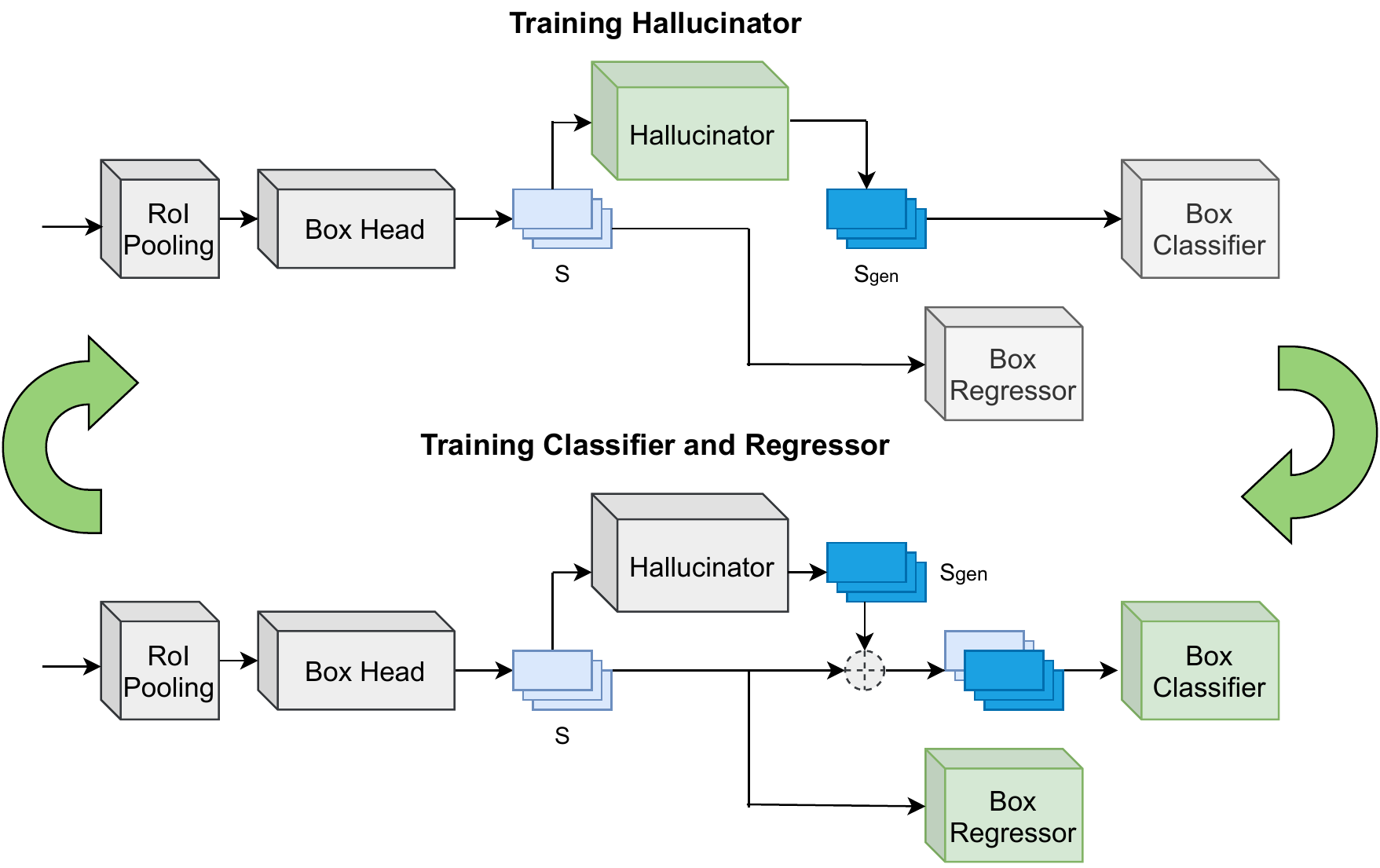}
\caption{Our EM-style training procedure for an RoI head with a hallucinator module. {\bf Upper:} how to train the hallucinator. The hallucinator is trained based on the classification loss on the hallucinated examples. All other modules are frozen when training the hallucinator. {\bf Lower}: how to train the classifier and the bounding box regressor. Both the original training examples and the hallucinated examples are used to train the classifier. Only the original training examples are used to train the bounding box regressor. The EM-style training {\em iterates} between the two processes. A {\color{darkslategray}gray} module indicates that it is frozen.}
\label{train_halluc}
\vspace{-4mm}
\end{figure}
As shown in Figure~\ref{halluc}, hallucination happens in the RoI head feature space. The hallucinator takes as input available training examples and generates hallucinated examples. The set of hallucinated examples $S_\text{gen}$ is then treated as additional training examples for learning the classifier on novel classes. Specially, given a seed example $x_i^{c_{k}}$ of category $c_{k}$, the hallucinator generates a hallucinated example by
\begin{equation}
\setlength{\abovedisplayskip}{3pt}
\setlength{\belowdisplayskip}{3pt}
\widetilde{x^{c_{k}}_i} = H(\mu_{c_{k}}, x_i^{c_{k}}, \varepsilon; \phi),
\end{equation}
where $\mu_{c_{k}}$ is the class prototype of $c_{k}$, which is the mean of all instances from $c_{k}$, and $\varepsilon$ is a per-example noise vector. Note that our hallucinator formulation is different from~\cite{wang2018low}: we introduce an additional input $\mu_{c_{k}}$ to explicitly capture the {\em global} category information. For $\widetilde{x^{c_{k}}_i}$, its label $y_{i}$ is the same as the seed example's category $c_{k}$. Assuming that we already have a trained box classifier, we directly use the classifier's (cross-entropy) classification loss on {\em all hallucinated examples} to train the hallucinator:

\begin{equation}
\setlength{\abovedisplayskip}{-10pt}
\setlength{\belowdisplayskip}{3pt}
L_\text{halluc} (\phi) = \sum_{i\in S_\text{gen}}^{} \left(-w^T_{y_{i}} \widetilde{x^{c_{k}}_i} +\log\sum_{k}^{} e^{w^T_{k} \widetilde{x^{c_k}_i}}\right),
\label{eq:hallluloss}
\end{equation}
where $w_{y_{i}}$ and $w_{k}$ are the {\em already-learned} classifier's weights on corresponding categories. This hallucination loss enforces the hallucinated examples to loosely ``agree with'' the trained classifier. In addition, since we take as input class prototypes, our hallucinator is implicitly regularized and thus not simply copying the seed examples. Compared with~\cite{wang2018low} which uses a complicated meta-learning process, we substantially simplify the training procedure.

\subsection{Training the Hallucinator and the Classifier}
As illustrated in Figure~\ref{train_halluc}, we propose an EM-style (expectation–maximization) training procedure to train the hallucinator and the detector's classifier: when training the hallucinator, the classifier is frozen; when training the classifier (and the box regressor), the hallucinator is frozen. Compared with end-to-end joint training, this iterative strategy improves the cooperation between the hallucinator and the classifier, thus producing examples that are more useful for building the few-shot classifier.

\textbf{Training on Base Classes:} In the stage of training on base classes, we first train the plain detector {\em without the hallucinator} in a standard way as before. We then insert the hallucinator into the detector, and freeze all model components except the hallucinator. Now, the hallucination will be performed in the pre-trained, fixed RoI head feature space. And the classifier is already trained on base classes using all available examples. We then use this base class classifier to guide the training of the hallucinator.

Note that, different from few-shot classification, the proposal generation process in detection produces several proposal boxes around an object, making a few training examples available even in the 1-shot scenario. We randomly sample these training examples as seed examples for the hallucinator. Consider a training batch $S$ which consists of $n$ base classes. For each class in the batch, the hallucinator generates $m$ examples, and $m$ is fixed for all batches. The set of hallucinated examples $S_\text{gen}$ is used to train the hallucinator based on the loss function~\eqref{eq:hallluloss}.

\textbf{Fine-Tuning on Novel Classes:} After training the hallucinator on base classes, we move on to the stage of fine-tuning the classifier and hallucinator on a few-shot dataset containing both base and novel classes. We freeze all model components prior to the hallucinator. Each training batch $S$ initially consists of an imbalanced set of foreground box examples $S_\text{pos}$ and background box examples $S_\text{neg}$, with background examples being the majority: $S = [S_\text{pos}; S_\text{neg}]$. The hallucinator generates a set of additional examples $S_\text{gen}$ for novel classes.
{\em Without changing the number of examples in total}, we randomly replace $|S_\text{gen}|$ background examples by $S_\text{gen}$ to obtain a refined training batch
\begin{equation}
\setlength{\abovedisplayskip}{3pt}
\setlength{\belowdisplayskip}{3pt}
\widetilde{S} = [S_\text{pos}, S_\text{gen}; S'_\text{neg}].
\end{equation} 
By doing so, we also partially alleviate the imbalance issue between foreground and background examples.
After training the classifier on the refined dataset with hallucinated examples, we fine-tune the hallucinator using the classifier based on Eq.~\eqref{eq:hallluloss}, then we use the fine-tuned hallucinator to fine-tune the classifier again, and so on. We stop this procedure after one or two iterations (\ie, the classifier is fine-tuned at most twice); empirically, we found that additional iterations do not further improve the performance. 


\section{Experiments}
\label{sec:exp}
\subsection{Implementation Details}
\textbf{Datasets and Evaluation Metrics:} We evaluate on two widely-used few-shot detection benchmarks: MS-COCO~\cite{lin2014microsoft} and PASCAL VOC (07 + 12)~\cite{everingham2010pascal}. For a fair comparison, the {\em same} base/novel category splits, train/test splits, and novel class instances~\cite{kang2018few, wang2020few} are used for training and evaluation. Consistent with recent work~\cite{kang2018few, wang2020few}, we report AP50 for three different base/novel splits under shots 1, 2, 3, 5, and 10 on PASCAL VOC; we report AP, AP50, and AP75 under shots 1, 2, and 3 on COCO. Note that we {\em particularly focus on the extremely-few-shot regime}.

\textbf{Baselines:} We mainly focus on the comparison with two state-of-the-art few-shot detectors -- TFA~\cite{wang2020few} and CoRPNs~\cite{CoRPNs}, and evaluate the effectiveness and generalizability of our hallucinator when combined with them. For both our models and the main baselines, we use Faster R-CNN~\cite{ren2015faster} as our base model. Following~\cite{wang2020few}, we use an ImageNet pre-trained ResNet-101 with a feature pyramid network~\cite{lin2017feature} as the backbone. All our experiments, including those with the main baselines, have ground-truth boxes appended as training examples in the RoI head. As reported in~\cite{wang2020few}, including ground-truth boxes leads to a 0.5\% AP gain on COCO. In addition, we compare with a variety of baselines, including concurrent work on few-shot detection. 

\begin{table*}[hbt!] 
\centering
\renewcommand{\arraystretch}{1.2} 
\resizebox{\textwidth}{!}{
\begin{tabular}{llccccc|ccccc|ccccc}
&& \multicolumn{5}{c}{Novel Set 1} & \multicolumn{5}{c}{Novel Set 2} & \multicolumn{5}{c}{Novel Set 3}\\
& Method & shot=1 & 2 & 3 & 5 & 10& shot=1 & 2 & 3 & 5 & 10 & shot=1 & 2 & 3 & 5 & 10\\
\hline
{Ours} & CoRPNs w/ cos + Halluc &\;\textcolor{blue}{\bf47.0} & {\bf44.9} & {\bf46.5} & {\bf54.7} & {54.7} & {\bf26.3} &{\bf31.8} & {\bf37.4} & {\bf37.4} & {41.2} & \textcolor{red}{\bf40.4} & \textcolor{red}{\bf42.1} & {43.3} & {51.4} & {49.6}\\
{Main baseline} & CoRPNs w/ cos~\cite{CoRPNs} &\;{44.4} & {38.5} & {46.4} & {54.1} & {\bf55.7} & {25.7} & {29.5} & {37.3} & {36.2} & {\bf41.3} & {35.8} & {41.8} & {\bf{44.6}} & \textcolor{blue}{\bf{51.6}} & {49.6}\\
\hline
{Ours} & TFA w/ cos + Halluc &\;{\bf45.1} & {\bf44.0} & {44.7} & {55.0} & {55.9} & {23.2} & {\bf27.5} & {\bf35.1} & {34.9} & {39.0} & {30.5} & {\bf35.1} & {41.4} & {49.0} & {49.3}\\
{Main baseline} & TFA w/ cos~\cite{wang2020few} &\;{39.8} & {36.1} & {44.7} & {\bf55.7} & {\bf56.0} & {\bf23.5} & {26.9} & {34.1} & {\bf35.1} & {\bf39.1} & {\bf30.8} & {34.8} & {\bf42.8} & {\bf49.5} & {\bf49.8}\\
\hline
& FRCN+ft-full~\cite{wang2020few} &\;{15.2} & {20.3} & {29.0} & {40.1} & {45.5} & {13.4} & {20.6} & {28.6} & {32.4} & {38.8} & {19.6} & {20.8} & {28.7} & {42.2} & {42.1}\\
& Meta R-CNN~\cite{yan2019metarcnn} &\;{19.9} & {25.5} & {35.0} & {45.7} & {51.5} & {10.4} & {19.4} & {29.6} & {34.8} & {45.4} & {14.3} & {18.2} & {27.5} & {41.2} & {48.1}\\
& CoAE*~\cite{Hsieh19AttenFew}&\;{12.7} & {14.6} & {14.8} & {18.2} & {21.7} & {4.4} & {11.3} & {20.5} & {18.0} & {19.0} & {6.3} & {7.6} & {9.5} & {15.0} & {19.0}\\
& MPSR~\cite{wu2020mpsr}&\;{41.7} & {43.1} & \textcolor{red}{51.4} & {55.2} & \textcolor{blue}{61.8} & {24.4} & {29.5} & {39.2} & {39.9} & {47.8} & {35.6} & {40.6} & {42.3} & {48.0} & {49.7}\\
{Other baselines} & FsDetView~\cite{Xiao2020FSDetView}&\;{24.2} & {35.3} & {42.2} & {49.1} & {57.4} & {21.6} & {24.6} & {31.9} & {37.0} & {45.7} & {21.2} & {30.0} & {37.2} & {43.8} & {49.6}\\
& NP-RepMet \cite{yang2020restoring}&\;{37.8} & {40.3} & {41.7} & {47.3} & {49.4} & \textcolor{red}{41.6} & \textcolor{red}{43.0} & \textcolor{blue}{43.4} & \textcolor{red}{47.4} & \textcolor{blue}{49.1} & {33.3} & {38.0} & {39.8} & {41.5} & {44.8}\\
& FSCE \cite{sun2021fsce}&\;{44.2} & {43.8} & \textcolor{red}{51.4} & \textcolor{red}{61.9} & \textcolor{red}{63.4} & {27.3} & {29.5} & \textcolor{red}{43.5} & \textcolor{blue}{44.2} & \textcolor{red}{50.2} & {37.2} & \textcolor{blue}{41.9} & \textcolor{red}{47.5} & \textcolor{red}{54.6} & \textcolor{red}{58.5} \\
& CME \cite{li2021maxmargin}&\;{41.5} & \textcolor{blue}{47.5} & {50.4} & \textcolor{blue}{58.2} & {60.9} & {27.2} & {30.2} & {41.4} & {42.5} & {46.8} & {34.3} & {39.6} & \textcolor{blue}{45.1} & {48.3} & \textcolor{blue}{51.5} \\
& SRR-FSD \cite{zhu2021semantic}&\;\textcolor{red}{47.8} & \textcolor{red}{50.5} & {51.3} & {55.2} & {56.8} & \textcolor{blue}{32.5} & \textcolor{blue}{35.3} & {39.1} & {40.8} & {43.8} & \textcolor{blue}{40.1} & {41.5} & {44.3} & {46.9} & {46.4} \\\end{tabular}
}
\vspace{0.1cm}
\caption{Few-shot detection performance (AP50) on PASCAL VOC novel classes under three base/novel splits. We follow the standard evaluation procedure in~\cite{wang2020few}. {\em The main comparison} focuses on combining hallucination (denoted as  `Halluc') with two state-of-the-art few-shot detectors: TFA and CoRPNs with cosine classifier (denoted as `w/ cos'). All models build upon Faster R-CNN with ResNet-101 backbone. `CoRPNs + Halluc' and CoRPNs {\em share the same hyper-parameter settings}; and the RPN outputs in `CoRPNs + Halluc' are the same as CoRPNs. Similarly, `TFA + Halluc' and TFA share the same RPN outputs. Hallucination yields significant improvements in the extremely-few-shot regime (1-shot and 2-shot). In higher shots, hallucination maintains comparable results to its baselines. In addition, we compare with other state-of-the-art methods, including some concurrent work. *Model re-evaluated under the standard procedure. Results in \textcolor{red}{red} are the best, results in \textcolor{blue}{blue} are the second-best, and results in {\bf{bold}} are the better ones in comparison with the corresponding baseline.}
\vspace{-4mm}
\label{tab:voc_novel}
\end{table*}

\textbf{Evaluation Procedure:} Both our models and the main baselines are $(|C_\text{b}|+|C_\text{n}|)$-way few-shot detectors. And we evaluate them on both base and novel classes under the standard evaluation procedure~\cite{wang2020few}, as described in Section~\ref{sec:method}. Some other baselines like~\cite{Hsieh19AttenFew} were initially evaluated under different procedures. For a fair comparison, {\em all reported numbers for those methods are the re-evaluated results under the standard evaluation procedure}.

Note that, following~\cite{wang2020few}, the fine-tuning stages on PASCAL VOC and COCO are slightly different. On PASCAL VOC, the classifier's weights on novel classes are randomly initialized and directly trained using a balanced dataset with base and novel classes. On COCO, we first train a $|C_\text{n}|$-way classifier on novel class instances. The trained classifier is used as an initialization of the classifier's weights on novel classes. We then train a $(|C_\text{b}|+|C_\text{n}|)$-way classifier using a balanced few-shot dataset with base and novel classes.

\textbf{Hallucinator Architecture:} We use a two-layer MLP (multi-layer perceptron) with ReLU as the hallucinator. Given that the inputs to the hallucinator are a class prototype, a seed example, and random noise, the input size is three times the feature size; the output size of each linear layer is the same size as the input feature. The seed example is randomly sampled from all examples in a training batch. The input noise is dataset-dependent. For each dataset, we calculate the mean and standard deviation of pre-trained input features and use those to sample the input noise. 

We pre-compute base class prototypes using all available examples before training the hallucinator. During training the hallucinator, we do not update these base class prototypes. In a different manner, we construct novel class prototypes {\em dynamically}, when using the learned hallucinator to train classifiers on novel classes. Specifically, novel class prototypes are computed by using both the training and hallucinated examples (in contrast to that only training examples are used for base class prototypes). For a novel class, whenever a new training example comes in or a new example is generated, its corresponding prototype will be updated. Despite somewhat inconsistency between base and novel classes for constructing prototypes, we found that this strategy exploits all available examples and empirically achieves the best performance.

\textbf{Hallucinator Initialization:} Following a similar strategy as in~\cite{wang2018low}, we initialize the hallucinator by using block diagonal identity matrices plus small random noise. The initialization noise is sampled from a normal distribution with a zero mean and a standard deviation of 0.02. As a form of regularization, this identity initialization ensures that the initially hallucinated examples are not too far away from the seed examples and thus do not degrade the performance.

\textbf{Hyper-Parameter Settings:} On PASCAL VOC, we train the hallucinator with batch size 16 and learning rate 0.02 for 8,000 iterations. We decay the learning rate by ratio 0.1 at 2,000 and 6,000 iterations.  On COCO, we train the hallucinator with batch size 64 and learning rate 0.02 for 21,200 iterations. We decay the learning rate by ratio 0.1 at 6,400 and 19,200 iterations.

\textbf{Number of Hallucinated Examples:} When training the classifier on novel classes, the hallucinator produces a fixed number ($m$) of examples per category in each training batch. For simplicity, we set $m$ to be the same as the average number of per-class training examples (\ie, RoI features), averaged over the training batches. On PASCAL VOC, there are roughly 20 training examples per class (with batch size 16), and we thus hallucinate 20 examples per class accordingly {\em for all experiments}. We keep this number on COCO as well. In the ablation study, we show that by tuning the number of hallucinated examples, the performance gets further improved. Note that this number should change when the number of images per batch varies.

\subsection{Main Results}
\textbf{Comparison with Main Baselines TFA and CoRPNs:} Tables~\ref{tab:voc_novel} and~\ref{tab:coco_novel} summarize the results for novel classes on PASCAL VOC and COCO, respectively. We have the following key observations. 

\begin{table*}[hbt!]
\centering
\renewcommand{\arraystretch}{1.1} 
\resizebox{0.83\textwidth}{!}{
\begin{tabular}{llccc|ccc|ccc}
&&\multicolumn{3}{c}{1-shot} & \multicolumn{3}{c}{2-shot} & \multicolumn{3}{c}{3-shot} \\
& Method &\; AP & AP50 & AP75 & AP & AP50 & AP75 & AP & AP50 & AP75 \\
\hline
Ours & CoRPNs w/ cos + Halluc &\;\textcolor{red}{\bf4.4} & \textcolor{blue}{\bf7.5} & \textcolor{red}{\bf4.9} & \textcolor{red}{\bf5.6} & \textcolor{blue}{\bf9.9} & \textcolor{red}{\bf5.9} & \textcolor{red}{\bf7.2} & \textcolor{blue}{\bf13.3} & \textcolor{red}{\bf7.4} \\
Main baseline & CoRPNs w/ cos~\cite{CoRPNs} &\;\textcolor{blue}{4.1} & {7.2} & \textcolor{blue}{4.4} & \textcolor{blue}{5.4} & {9.6} & \textcolor{blue}{5.6} & \textcolor{blue}{7.1} & {13.2} & \textcolor{blue}{7.2} \\
\hline
Ours & TFA w/ cos + Halluc&\;{\bf3.8} & {\bf6.5} & {\bf4.3} & {\bf5.0} & {\bf9.0} & {\bf5.2} & {\bf6.9} & {\bf12.6} & {\bf7.0} \\
Main baseline & TFA w/ cos~\cite{wang2020few} &\;{3.4} & {5.8} & {3.8} & {4.6} & {8.3} & {4.8} & {6.6} & {12.1} & {6.5} \\
\hline
\multirow{2}{*}{Other baselines} & MPSR**~\cite{wu2020mpsr} &\;{2.3} & {4.1} & {2.3} & {3.5} & {6.3} & {3.4} & {5.2} & {9.5} & {5.1}\\
& FsDetView~\cite{Xiao2020FSDetView} &\;{3.2} & \textcolor{red}{
\bf8.9} & {1.4} & {4.9} & \textcolor{red}{\bf13.3} & {2.3} & {6.7} & \textcolor{red}{\bf18.6} & {2.9}\\
\end{tabular}
}
\vspace{0.1cm}
\caption{Few-shot detection performance on COCO novel classes in 1, 2, and 3-shot scenarios. See the caption of Table~\ref{tab:voc_novel} for details. Hallucination {\em consistently improves both baselines in all settings}. `CoRPNs + Halluc' achieves the state-of-the-art results in most cases. **Model evaluated using the public code and the pre-trained detector on base classes.} 
\vspace{-4mm}
\label{tab:coco_novel}
\end{table*}

\begin{table*}[hbt!] 
\centering
\renewcommand{\arraystretch}{1.2} 
\resizebox{0.86\textwidth}{!}{
\begin{tabular}{llccc|ccc|ccc}
&&\multicolumn{3}{c}{1-shot fine-tuned} & \multicolumn{3}{c}{2-shot fine-tuned} & \multicolumn{3}{c}{3-shot fine-tuned} \\
& Method &\; AP & AP50 & AP75 & AP & AP50 & AP75 & AP & AP50 & AP75 \\
\hline
{Ours} & CoRPNs w/ cos + Halluc &\;{32.3} & {52.4} & {34.4} & {34.5} & {55.3} & {37.0} & {34.7} & {55.1} & {37.5}\\
{Main baseline} & CoRPNs w/ cos~\cite{CoRPNs}  &\;{34.1} & {55.1} & {36.5} & {34.7} & {55.3} & {37.3} & {34.8} & {55.2} & {37.6}\\
\hline
{Ours} & TFA w/ cos + Halluc &\;{31.5} & {50.8} & {33.9} & {32.8} & {52.4} & {35.4} & {33.3} & {52.8} & {36.4}\\
{Main baseline} & TFA w/ cos~\cite{wang2020few}  &\;{34.1} & {54.7} & {36.4} & {34.7} & {55.1} & {37.6} & {34.7} & {54.8} & {37.9}\\
\hline
\multirow{2}{*}{Other baselines} & MPSR**~\cite{wu2020mpsr} &\;{12.1} & {17.1} & {14.2} & {14.4} & {20.7} & {16.9} & {15.8} & {23.3} & {18.3}\\
& FsDetView~\cite{Xiao2020FSDetView} &\;{2.4} & {7.0} & {1.0} & {4.4} & {11.9} & {2.2} & {4.9} & {13.6} & {2.2}\\
\end{tabular}
}
\vspace{0.1cm}
\caption{Detection performance on COCO {\em base} classes, {\em after fine-tuning} with a balanced dataset (1, 2, and 3-shots) containing base and novel class instances. Hallucination slightly degrades base class performance compared with CoRPNs and TFA, but it substantially outperforms state-of-the-art MPSR and FsDetView. **Model evaluated using the public code and the pre-trained detector on base classes.}
\vspace{-4mm}
\label{tab:coco_base}
\end{table*}

(1) Hallucination improves the performance over the main baselines {\em by large margins in the extremely-few-shot regime}. On PASCAL VOC, hallucination yields substantial improvements in 1-shot and 2-shot scenarios. On the more challenging COCO benchmark, hallucination {\em consistently} improves performance when combined with both baselines.

(2) Our hallucination strategy is {\em general and applicable to different types of detectors regardless of RPN outputs}. This is because hallucination provides additional, useful sample variation that is independent of RPN outputs. 

(3) On PASCAL VOC (Table~\ref{tab:voc_novel}), hallucination demonstrates slightly different behaviors when combined with CoRPNs or TFA: while hallucination consistently improves CoRPNs in the extremely-few-shot regime, it is not always helpful for TFA in some cases (though hallucination also does not hurt). A possible explanation is that our hallucinator tends to be {\em conservative}, making the hallucinated examples still close to the original seed training examples. Hence, in the cases (\eg, novel sets 2 and 3) where the training and test examples are quite distinct, the hallucinated examples produced from the training examples could not help build a better classifier for detecting the test examples. In such cases, a more {\em aggressive} hallucinator is desired to generate examples that are further away from the seed examples. The ablation study in Section~\ref{alter_option} investigates this issue in more depth and supports our observation.

(4) The performance gain of hallucination diminishes as the number of training examples increases. In higher-shot scenarios, our current hallucination does not bring in additional within-class variation beyond the given training examples, and is thus no longer beneficial. Note that, however, hallucination does not hurt the performance in 5/10-shot scenarios -- performance variation is within 95\% confidence interval over multiple random samples according to~\cite{wang2020few}. Given that a {\em single} hallucinator is trained in our case, we might need different hallucinators or hallucination strategies in different sample size regimes.

\textbf{Comparison with Other State-of-the-Art Methods:} Tables~\ref{tab:voc_novel} and~\ref{tab:coco_novel} show that our approach is superior or comparable to other few-shot detection methods. In particular, we achieve very competitive results in the extremely-few-shot regime. We believe that our effort is orthogonal to concurrent work; our hallucination strategy can be combined with other methods to further improve their performance.

\textbf{Base Class Performance:}
Table~\ref{tab:coco_base} presents the detection results on base classes, evaluated after fine-tuning with novel class instances. We observe that hallucination improves novel class performance at the expense of {\em slight} degradation of performance on base classes. This issue could be potentially addressed by using a smaller fine-tuning learning rate for the classifier components of base classes (which have been already well-trained). More notably, compared with other state-of-the-art methods such as MPSR~\cite{wu2020mpsr} and FsDetView~\cite{Xiao2020FSDetView}, we achieve a significantly better trade-off between novel and base class performance.
\subsection{Ablation Studies}
\label{sec:ablation}
\textbf{EM vs. Joint Training Procedures:} Table~\ref{tab:EM} shows that, with the same hallucinator architecture, our EM-style training procedure significantly outperforms joint training, indicating that joint training might be greedy in this case. In addition, 2 EM-iterations outperform 1 iteration, and we found that more iterations do not improve the performance.
\begin{table}
\centering
\renewcommand{\arraystretch}{1.1}
\resizebox{.88\columnwidth}{!}{%
\begin{tabular}{l|c|c|c}
{Method} & {Novel Set 1} & {Novel Set 2} & {Novel Set 3}  \\
    \hline
    {Joint training} &  38.5 & 22.2 & 33.6\\
    {EM w/ 1 iter} &  46.7 & 24.5 & 38.5\\
    {EM w/ 2 iter} & {\bf47.0} & {\bf26.3} & {\bf40.4}\\
  \end{tabular}
  }
 \vspace{0.05cm}
 \caption{Comparison (1-shot AP50) of different training procedures for CoRPNs with hallucination on PASCAL VOC. All procedures use {\em the same hallucinator architecture}. {\em EM-style training significantly outperforms joint training} under all base/novel splits. The second EM iteration also {\em consistently improves performance}.}
\label{tab:EM}
\end{table}

\textbf{Conservative vs. Aggressive Hallucinators:}
\label{alter_option}
In the main experiments, we show that a conservative hallucinator is not always helpful and in some cases, we need a more aggressive hallucinator. Here we build such an aggressive hallucinator by {\em changing the hallucination space}. As shown in Figure~\ref{halluc}, the original hallucinator generates examples {\em directly in the operational space of the classifier} (\ie, in the feature space after the box head and right before the classifier); this makes it conservative and reluctant to produce the kinds of examples that drastically change the classifier decision boundaries. By contrast, our new hallucinator generates examples in the feature space {\em before the box head} and accordingly, it becomes a three-layer convolutional network. Now, the box head makes allowances for errors in the hallucination, and the hallucinator becomes more aggressive and allows radical change of classifiers. We also make some additional modifications as follows: (1) we jointly train the aggressive hallucinator with other model components, and (2) we use a cosine prototypical network loss~\cite{snell2017prototypical} computed on held-out validation examples with hallucinated prototypes as additional guidance to train the hallucinator. Table~\ref{tab:joint_option} shows that the aggressive hallucinator significantly helps the cases where the performance of the conservative hallucinator is lagging. Importantly, neither of the two hallucinator variants hurts the performance.

\begin{table}
\centering
\renewcommand{\arraystretch}{1.1}
\resizebox{\columnwidth}{!}{%
\begin{tabular}{l|c|c|c}
{Method} & {Novel Set 1} & {Novel Set 2} & {Novel Set 3}  \\
    \hline
    {TFA~\cite{wang2020few}} &  39.8 & 23.5 & 30.8\\
    {TFA + Halluc, Conservative} &  {\bf45.1} & {23.2} & {30.5}\\
    {TFA + Halluc, Aggressive} &  {40.8} & {\bf29.3} & {\bf33.7}\\
  \end{tabular}
  }
 \vspace{0.05cm}
 \caption{Analysis of conservative and aggressive hallucinators: 1-shot AP50 on PASCAL VOC under three base/novel splits. The aggressive hallucinator brings substantial improvements over TFA on novel sets 2 and 3 without hurting the performance on novel set 1. The conservative hallucinator significantly outperforms TFA on novel set 1, and performs comparably on the other two novel sets.}
\label{tab:joint_option}
\end{table}

\textbf{Number of Hallucinated Examples:} 
Our hallucinator generates a fixed number of examples per class in each training batch (20 examples in the main results). Table~\ref{tab:halluc_num} investigates the impact of the number of hallucinated examples. As it increases from 0 to 20, the performance gradually improves and then saturates and drops slightly. Note that the drop is still within 95\% confidence interval over multiple random samples according to~\cite{wang2020few}. This is because even if we produce a small number of hallucinated examples per training batch (\eg, 3), after many iterations, the {\em cumulative} number of hallucinated examples is large enough.


\begin{table}
\centering
  \resizebox{0.58\columnwidth}{!}{%
\begin{tabular}{l|c|c}
{Method} & {\#Halluc}  & {AP50}  \\
    \hline
    {CoRPNs~\cite{CoRPNs}} & {0} & {44.4} \\
    \hline
    \multirow{6}{*}{CoRPNs + Halluc} &  {1} & {47.1} \\
     &  {2} & {47.0} \\
     &  {3} & {\bf48.0} \\
     & {5} & {47.9} \\
     & {10} & {47.8} \\
     & {20} & {47.0} \\
  \end{tabular}
  }
 \vspace{0.05cm}
 \caption{Impact of the number of hallucinated examples on 1-shot detection (AP50) under PASCAL VOC novel set 1.}
\label{tab:halluc_num}
\end{table}
\textbf{Results with Fully-Connected Classifiers: }
We use cosine classifiers in the main results. Table~\ref{tab:fc} shows detection results using fully-connected classifiers. Hallucination improves CoRPNs in all 1-shot cases, suggesting the {\em effectiveness of hallucination irrespective of the classifier choice}.
\begin{table}
\centering
  \resizebox{\columnwidth}{!}{%
\begin{tabular}{l|c|c|c}
{Method} & {Novel Set 1}  & {Novel Set 2} & {Novel Set 3} \\
    \hline
    {CoRPNs w/ fc ~\cite{CoRPNs}} & {40.8} & {20.4} & {29.4}\\
    {CoRPNs w/ fc + Halluc} & {\bf44.2} & {\bf23.0} & {\bf31.5} \\
  \end{tabular}
  }
 \vspace{0.1cm}
 \caption{1-shot detection performance (AP50) on PACAL VOC novel classes with fully-connected classifiers (`w/ fc'). CoRPNs and `CoRPNs + Halluc' use the same set of hyper-parameters and share the same RPN outputs. `CoRPNs + Halluc' outperforms CoRPNs under all three base/novel splits.}
\label{tab:fc}
\end{table}

\textbf{Performance Gain Analysis: }
We investigate where our performance gain comes from. The detection AP can be improved in two ways: (1) more boxes are correctly identified, \ie, true positive rate goes up; (2) fewer boxes are misclassified, \ie, false positive rate goes down. As shown in Table~\ref{tab:analysis}, hallucination {\em both} increases the number of true positives in most cases and always decreases the number of false positives. We find that the scale of improvements on false positives is larger than the scale of improvements on true positives. A possible explanation is that hallucination generates {\em close-to-boundary} examples, and {\em improves decision boundaries}. Many misclassified boxes are eliminated, since we have a {\em better estimation of decision boundaries}.



\begin{table}
\centering
  \resizebox{.83\columnwidth}{!}{%
  \renewcommand{\arraystretch}{1.1}
\begin{tabular}{lc|c}
\multirow{2}{*}{Novel Set 1} & {Avg. True Positive} & {Avg. False Positive}\\
 & {Boxes ($\uparrow$)} & {Boxes ($\downarrow$)}\\
    \hline
    {TFA~\cite{wang2020few}} &  118.9 & 4096.6\\
    {TFA + Halluc} & \bf{132.0} & \bf{3309.3}\\
    \hline
    {CoRPNs~\cite{CoRPNs}} &  133.9 & 3440.6\\
    {CoRPNs + Halluc} & \bf{137.0} & \bf{3041.1}\\
\hline
\multicolumn{3}{l}{Novel Set 2} \\
    \hline
    {TFA~\cite{wang2020few}} &{\bf76.8} & 897.1\\
    {TFA + Halluc} & 69.5 & {\bf844.9}\\
    \hline
    {CoRPNs~\cite{CoRPNs}} & {\bf78.7} & 920.1\\
    {CoRPNs + Halluc} & 77.1 & {\bf826.6}\\
\hline
\multicolumn{3}{l}{Novel Set 3}\\
    \hline
    {TFA~\cite{wang2020few}} &  {109.3} & {2043.2}\\
    {TFA + Halluc} & {\bf113.6} & {\bf1461.4}\\
    \hline
    {CoRPNs~\cite{CoRPNs}} &  {110.3} & {1704.7}\\
    {CoRPNs + Halluc} & {\bf117.6} & {\bf1345.0}\\
  \end{tabular} 
  }
 \vspace{0.12cm}
 \caption{Average total number of true positive boxes and false positive boxes in novel classes for different 1-shot models on PASCAL VOC. Hallucination increases the number of true positive boxes in most cases and decreases the number of false positive boxes, with a larger improvement in the false positive rate. }
\label{tab:analysis}
\end{table}

\section{Conclusion and Future Work}
\label{sec:conclu}
This paper tackles the lack of sample variation problem in few-shot detection. We introduce a hallucinator network to generate additional feature-level training examples. Experimental evaluation demonstrates that data hallucination, as a general strategy, can be incorporated into different types of detectors and substantially improves few-shot detection. Our analysis further shows that different design choices will change the behavior of the hallucinator.

Our method is one possible instantiation of the data hallucination strategy. We believe that further refinements of the proposed hallucinator architectures and exploration of a more dedicated control of training procedure could lead to a more effective hallucinator with better performance. Finally, it is a promising avenue to continue this line of work for other visual recognition tasks beyond object detection.

\clearpage
{\small
\bibliographystyle{ieee_fullname}
\bibliography{egbib}
}
\end{document}